\documentclass[letterpaper,twocolumn,fleqn]{article} 
\usepackage{ist}
\usepackage{authblk}
\usepackage{graphicx,subfigure}
\usepackage{url}

\title{Towards the Creation of a Nutrition \\ and Food Group Based Image Database}
\author[1]{Zeman Shao}
\author[1]{Jiangpeng He}
\author[1]{Ya-Yuan Yu}
\author[2]{Luotao Lin}
\author[2]{Alexandra E Cowan}
\author[2]{Heather A Eicher-Miller}
\author[1]{Fengqing Zhu}
\affil[1]{Elmore Family School of Electrical and Computer Engineering, Purdue University, West Lafayette, Indiana, USA}
\affil[2]{Department of Nutrition Science, Purdue University, West Lafayette, Indiana, USA}

\date{} % date has an empty field.

\begin{document} 

\maketitle 

\thispagestyle{empty} % prevents the first page to be numbered

%%%%%%%%%%%%%%%%%%%%%%%%%%%%%%%%%%
% Abstract
%%%%%%%%%%%%%%%%%%%%%%%%%%%%%%%%%%

\begin{abstract}
Food classification is critical to the analysis of nutrients comprising foods reported in dietary assessment. 
% promotes a informed food decision for a healthy diet.
Advances in mobile and wearable sensors, combined with new image based methods, particularly deep learning based approaches, have shown great promise to improve the accuracy of food classification to assess dietary intake. 
% Emerging deep learning based approaches haven shown great improvements on food image analysis. 
However, these approaches are data-hungry and their performances are heavily reliant on the quantity and quality of the available datasets for training the food classification model. Existing food image datasets are not suitable for fine-grained food classification and the following nutrition analysis as they lack fine-grained and transparently derived food group based identification which are often provided by trained dietitians with expert domain knowledge.
In this paper, we propose a framework to create a nutrition and food group based image database that contains both visual and hierarchical food categorization information to enhance links to the nutrient profile of each food.
We design a protocol for linking food group based food codes in the U.S. Department of Agriculture's (USDA) Food and Nutrient Database for Dietary Studies (FNDDS) to a food image dataset, and implement a web-based annotation tool for efficient deployment of this protocol.
% To improve the efficiency of the linking process, we implement a web-based annotation tool for nutrition researchers to match each food item location represented by the bounding box on the image to the associated nutrition food code.
Our proposed method is used to build a nutrition and food group based image database including 16,114 food images representing the 74 most frequently consumed What We Eat in America (WWEIA) food sub-categories in the United States with 1,865 USDA food code matched to a nutrient database, the USDA FNDDS nutrient database. 
% which is valuable for future research for image-based nutrition analysis.

\end{abstract}

%%%%%%%%%%%%%%%%%%%%%%%%%%%%%%%%%%%%
% Introduction
%%%%%%%%%%%%%%%%%%%%%%%%%%%%%%%%%%%%
\section{Introduction}
\label{sec:intro}

Food image analysis is an important research problem for many health applications such as the analysis of nutrients, dietary assessment, diet management, nutrition intervention and food recommendation~\cite{zhu2010A, six2010, shao2021integrated, Meyers_2015_ICCV, kong2012, joutou2009, foodlogA}.
Recently, deep learning based methods have been developed for food recognition and portion estimation~\cite{Meyers_2015_ICCV, He_2021_ICCVW, yanai_2017,he2020multi,he2020multi,tada_segmentation,wang-icip2017, he2021end, Mao2021,Fang2018,shao2021towards}, showing a great potential for advancing the field of food image analysis. 
We are interested in developing deep learning methods for image-based food classification from both visual and food group based information.
% which promotes a informed food decision for a healthy diet.
However, most deep learning based methods are data-hungry and their performances are heavily reliant on the quantity and quality of the available datasets for training and to validate the deep learning models. 
% There still exists a significant obstacle for providing nutrition information in this research field of deep learning based nutrition analysis.
There are a lack of annotated food datasets with accurate classification of foods based on a hierarchical food categorization system, embedding individual foods within their respective sub-categories and general categories and linking to specific and broader nutrient information so that subsequent analysis of nutrient content can be carried out. 
The What We Eat in America (WWEIA) is such a system that provides a categorization for grouping similar foods and beverages consumed in the American diet together into WWEIA sub-categories and categories based on the nationally representative daily dietary intake information collected in the National Health and Nutrition Examination Survey~\cite{wweia}.

% and their membership to WWEIA food categories and sub-categories hinders further progress in image based food classification.

Although nutrient databases provide comprehensive nutrient information for each specific food that is also categorized into a respective WWEIA food category and sub-category, there are no food images associated with the specific food codes to match to image based dietary assessment methods and ultimately, perform image based nutrient analysis.
The U.S. Department of Agriculture's (USDA) Food and Nutrient Database for Dietary Studies (FNDDS) 2017-2018~\cite{fndds2015} is a federally maintained nutrient database that can be utilized to obtain nutrient profiles, including a panel of macronutrients and micronutrients, of foods and beverages.
Each food and beverage in the FNDDS database is linked to a specific USDA food code and food description, yet the FNDDS also lacks associated visual information for each food and beverage, and therefore, it cannot be utilized for image analysis task.

Several public food image datasets exist which mainly focus on the food recognition and portion estimation tasks but do not offer food group based classification along with the images.
Due to the lack of food group based information provided by these datasets, they cannot be used to accurately identify the food and match it to the corresponding food in a nutrient database like the FNDDS.
% cover the need of the image-based nutrition analysis task. 
Food image datasets, such as Food-101~\cite{bossard2014food} and UPMC-101~\cite{wang2015recipe}, are designed for food classification. Recipe image datasets, such as Recipe1M/Recipe1M+~\cite{salvador2017learning, marin2019recipe1m} and Vireo-172~\cite{chen2016deep} are designed for recipe retrieval. Note that they do not provide information on the location of the food in the image.
Although datasets such as UEC-100/UEC-256~\cite{matsuda2012recognition, kawano2014automatic}, UNIMIB2015/UNIMIB2016~\cite{ciocca2015food, ciocca2016food} and VFN~\cite{Mao2021} provide either bounding boxes or segmentation masks for each food item in the image and can be used for validating food localization methods, there is no food group or nutrition information to link the identified food to a nutrient database.
The Nutrition5k~\cite{thames2021nutrition5k} dataset contains groundtruth food labels and lists nutrient information such as protein, carb, fat, mass and calories for each food in the image, but details of how the nutrition information is obtained are unknown and there is no information about the location of the foods in the images. 
Overall, these datasets are not suitable for accurate food classification embedded in a system of WWEIA food categories or sub-categories that ultimately can be used to link the food to it's corresponding nutrient information within a nutrient database, since fine-grained and transparently derived dietary information or link to a nutrient database is not available.

In this paper, we propose a framework to create a food group based image database that contains nutrition information for each food in the images that is linked to a nutrient database, the FNDDS. 
We design a protocol for linking the food classification using USDA food codes to food images from a food image dataset. To reduce the burden of manually perform this linking process, we implement a web-based annotation tool for efficient deployment of this protocol.
% groundtruth information for validating food recognition method and also takes account of nutrition information.
% In addition to a protocol for linking visual information to the nutritional information, an annotation tool with a systematic design is required to further relieve the burden of nutrition researcher and speed up the linking process.
Most existing annotation tools, such as standalone annotation tools LabelMe~\cite{russell2008labelme} and the online crowdsourcing platform Amazon Mechanical Turk (AMT)~\cite{amtwebsite}, are designed and commonly used for general annotation purposes where one or multiple specific terms are labeled for a region in the image.
These tools are not suitable for collecting and identifying foods within a hierarchical system of food categorization which has a hierarchical data structure to be matched to the food image.
% which are not suitable for collecting nutrition information.
Leveraging a previously developed web-based annotation tool for collecting food images \cite{shao2019semi}, we further enhance this tool by introducing a feature to match the generic food labels to the USDA food codes in a systematic way.

% We have designed a web-based annotation tool to locate food items and assign generic food labels, which are visual based food terms not nutrition specific food term,  for online food image collected by web crawler\cite{shao2019semi}.
% We enhanced the tool with functionality to match the generic food labels to the USDA food codes in a systematic way.

The main contributions of this paper can be summarized as following:
\begin{itemize}
\item	Development of a food group based classification system to link the FNDDS database to the appropriate food images. 
\item	Creation of a web-based annotation tool that allows researchers to link food images to the appropriate USDA food codes in the FNDDS nutrient database.
\item	Development of a new food group based image database featuring the most frequently consumed foods by adults in the United States 
\end{itemize}

%%%%%%%%%%%%%%%%%%%%%%%%%%%%%%%%%%
% Method
%%%%%%%%%%%%%%%%%%%%%%%%%%%%%%%%%%

\section{Creation of a Nutrition and Food Group Based Image Database}
Food image analysis to accurately identify specific food or beverage items relies on visual information to predict the food type. However, accurate details of the food or beverage are required to further calculate the nutrient content of the food.
Both visual and nutrition information, such as items being whole-grain or low-fat, are considered when creating a nutrition and food group based image database, which is a challenging problem for several reasons.
Food items that are similar in nutritional content may differ in visual appearance, \textit{e.g.}, ``salmon" and ``cod". If both items are combined into a broad category like "fish" it could result in poor classification performance due to the large visual dissimilarity.
In contrast, food items that appear visually similar, \textit{e.g.}, ``meat loaf" and ``loaf of bread", may have very different nutritional content, and should be designated to their respective categories for accurate analysis of nutrients.
To address these challenges, we propose a framework to create a nutrition and food group based image database that considers both visual and food group based nutrient information jointly by following these steps:
\begin{enumerate}
    \item Determine the most commonly consumed WWEIA food sub-categories and those contributing most to total energy intake from the FNDDS and using the WWEIA food sub-categories, in order to more comprehensively capture the "most important" foods and food groups among the US population ~\cite{lin2022nhames}.
    \item Search and download food images from online websites using a web crawler, based on the selected WWEIA food sub-categories, followed by image cleaning and data refinement via automatic image analysis methods~\cite{shao2019semi}.
    \item Annotate the food images using an online crowdsourcing tool~\cite{fang2018ctada} with generic food labels and bounding boxes around the food items.
    \item Nutrition researchers review the annotated food images, their generic food labels and associated bounding boxes, and then link each food in a bounding box to a specific USDA food code. 
\end{enumerate}

\subsection{Identifying the "Most Important" Foods to Populate the Food Image Database, Step 1}
% Here we describe in detail the protocol for selecting the most frequently consumed WWEIA food sub-categories from the USDA WWEIA Database (Step 1) and linking visual information in the food image to the nutritional information through the food codes included in a nutrient database, \textit{i.e.}, the FNDDS database~\cite{fndds2015} (Step 4).
% Our goal is to develop a protocol to link food images to the food codes in the FNDDS database.

First, the National Health and Nutrition Examination Survey (NHANES) 2009-2016~\cite{nhanes} is used to determine the consumption frequency and energy contribution of each WWEIA food sub-categories~\cite{lin2022nhames}, which are used to create our image database.
The NHANES is a national survey that monitors the health and nutritional status of adults and children across the United States conducted by the National Center for Health Statistics of Centers for Disease Control and Prevention. 
The NHANES survey includes an in-person household interview and a health examination. 
NHANES participants' dietary recalls were collected during the physical health examination through the USDA Automated Multiple-Pass Method~\cite{ampm}. 
Participants reported the foods they consumed during the last 24 hours, including information on amount and type of each food, the time of intake, and a detailed food description~\cite{nhanes}. 
Each reported food item is linked with an 8-digit USDA food code from the USDA’s Food and Nutrient Database for Dietary Studies (FNDDS)~\cite{fndds5,fndds2011,fndds2013, fndds2015}. 
The digits in the USDA food codes have meaningful links to the hierarchical system of food sub-categories and categories so that they can also be used to sort the reported foods into the WWEIA food categories and sub-categories~\cite{wweia}.

% NHANES was used in~\cite{lin2022nhames}, to determine the consumption frequency and energy contribution of each WWEIA food category and sub-category, and food item that was reported by participants from NHANES, and ranked the WWEIA food category and sub-category from high consumption frequency to low and higher energy contribution to low. 
Following the protocol described in~\cite{lin2022nhames}, we chose the most frequently consumed food sub-categories from WWEIA and those that also contribute the most to total energy intake in the population of US adults. In some cases, a specific food code is assigned to each food in the image. In cases where the specific food code cannot be determined from the image, we assigned a general USDA food code, \textit{i.e.} ``NFS (Not Further Specified)". Each image in our image database is reviewed by at least 3 nutrition researchers to ensure consistent assignment of the corresponding food codes. 
An example of the USDA food codes for ``Almond" is shown in Table~\ref{tab:fndds}, where ``almonds, NFS (not further specified)" is assigned as the general USDA food code description with corresponding USDA food code 42100100 assigned to most images.
% Then nutrition researchers reviewed each image and linked the foods in the image to the specific USDA food code. 
% For example of the ``Almond", the general USDA food code 42100100 with its corresponding description ``almonds, NFS" is assigned to most images. 
When other characteristics of the food are visible in the images such as seasonings, a more specific USDA food code is used, such as ``Almonds, flavored” with USDA food code 42101300. However, it is important to note that specific characteristics of the food are not assigned beyond what is truly observable from the image, such as "Almonds, salted" when only an unknown seasoning is apparent as these designations impact the nutrient composition that the USDA food code links to in the FNDDS database.
% What We Eat in America (WWEIA)~\cite{wweia} is the dietary intake interview component of the National Health and Nutrition Examination Survey (NHANES)~\cite{nhanes} and also shows the intake frequency of each of the WWEIA food categories during the interviews.
% FNDDS~\cite{fndds2018} provides the energy and nutrient values for foods and beverages reported in WWEIA.
% The nutrition researcher first determine a list of food codes from FNDDS database that correspond to all food categories selected from WWEIA database.
% Next, a general USDA food code is selected for each food category from WWEIA database,
% that can be apply to all images in a specific food category assigned to the relevant images.

% One of the most important reference to determine the general USDA food code is the consumption frequency and energy contribution rankings from the NHANES survey~\cite{nhanes}.  The NHANES survey provides a list of USDA food codes that are commonly consumed and contribute the most to energy among different population.
% Nutrition researchers review each image and classify the image to the most specific USDA food code possible, whether it be a food category or subcategory.
\begin{table}[t]
    \caption{Table~\ref{tab:fndds}: Example list of food codes for the ``Almond" category from the frequency and energy contribution ranking list. In this case, the ``almonds, Not Further Specified (NFS)" is the most commonly consumed food code for the ``Almond” category}
    \label{tab:fndds}
    \begin{center} 
    \resizebox{0.4\textwidth}{!}{
    \begin{tabular}{ |p{0.19\columnwidth}|p{0.38\columnwidth}| p{0.3\columnwidth}|} 
     \hline
     Food Code & Main Food Description & WWEIA Category Description \\
     \hline
     42100100 & Almonds, NFS &  Nuts and seeds \\ 
     \hline
     42101000 & Almonds, unroasted &  Nuts and seeds \\ 
     \hline
     42101110 & Almonds, salted &  Nuts and seeds \\ 
    %  \hline
    %  42101120 & Almonds, lightly salted & Nuts and seeds \\ 
    %  \hline
    %  42101130 & Almonds, unsalted &  Nuts and seeds \\ 
     \hline
    \end{tabular}
    }
    \end{center}
    \vspace{-1.0cm}
\end{table}
\begin{figure}[t]
    \centering
    \begin{subfigure}
    \centering
    \includegraphics[width=0.3\columnwidth]{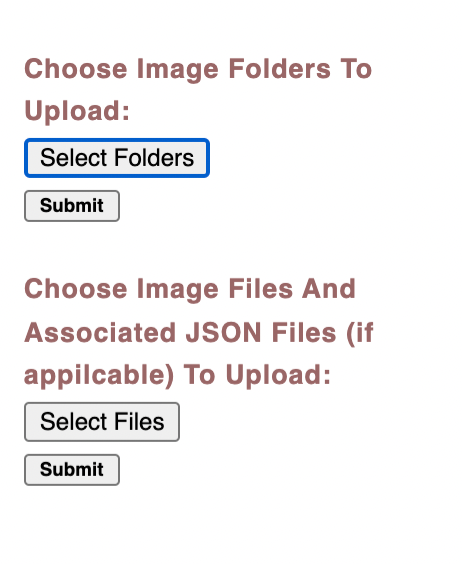}
    \caption{Screenshot of importing images into the annotation tool}
    \label{fig:import}
    \end{subfigure}
    
    \begin{subfigure}
    \centering
    \includegraphics[width=0.6\columnwidth]{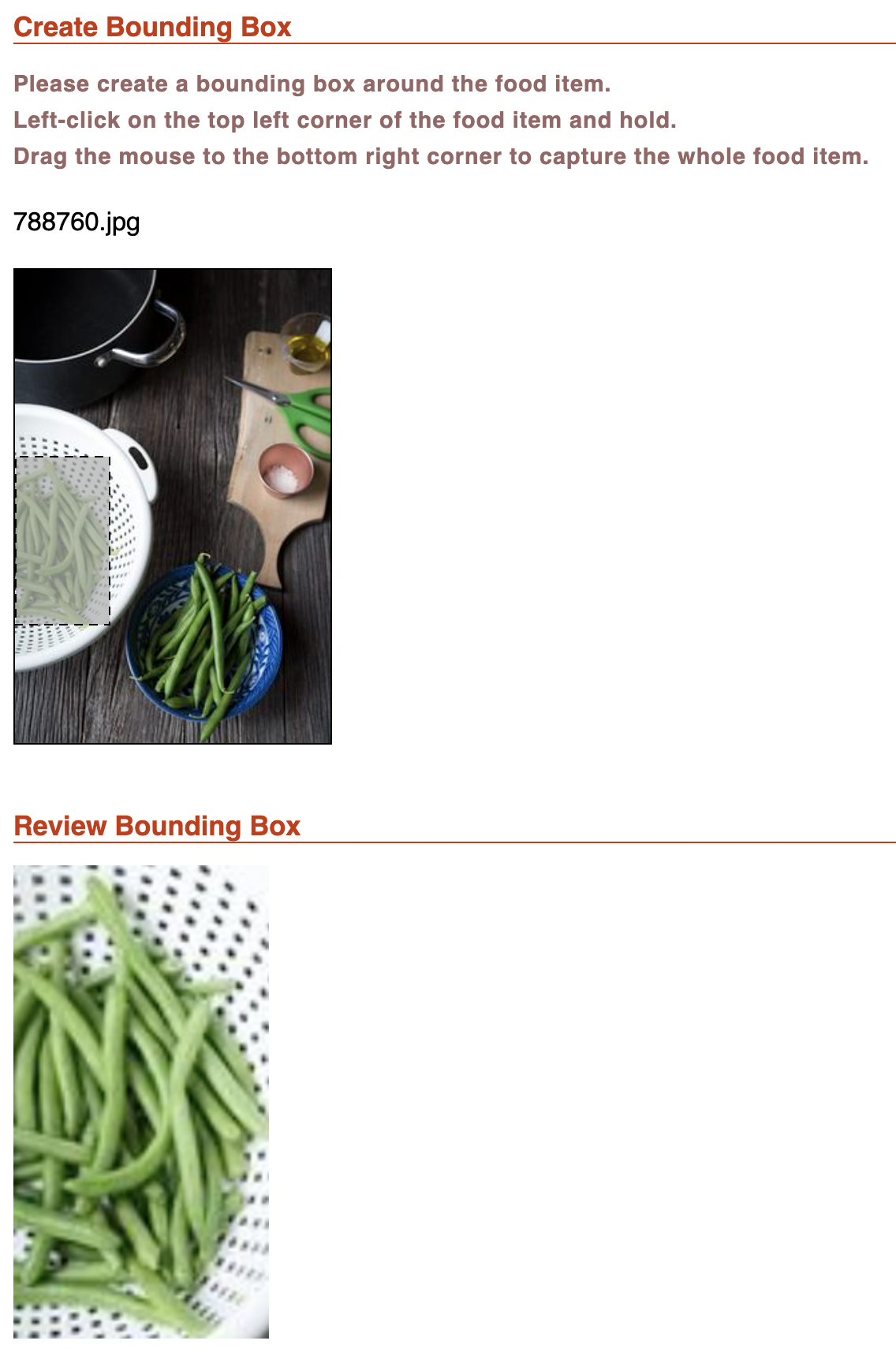}
    \caption{Example of localizing the food item on the image}
    \label{fig:loc}
    \end{subfigure}
\end{figure}

\subsection{Populate, Annotate, and Review Images, Step 2-4}
To relieve the burden of annotator and speed up the data curation process, we previously designed and implemented a semi-automatic system for online food image collection and annotation, including a web crawler, automatic food detection and online crowdsourcing tool~\cite{shao2019semi}.
We use the web crawler to download large sets of online food images based on the selected WWEIA food sub-category, followed by image cleaning and data refinement via automatic food detection (Step 2).

Although the online crowdsourcing tool provides an efficient way to remove noisy image and annotate the food location on the image (Step 3), it does not have the capability to annotate group based identification where one or more specific USDA food code is matched to each food item in an image. 
Therefore, our goal is to enhance the annotation tool by implementing features to link food images to USDA food codes (Step 4). 
We introduce four new features in the web-based annotation tool to accomplish this goal. 

\subsubsection{Image Importing}
We designed two options for the nutrition researchers to load images into the tool, one for the image folder and another one for the image files, as shown in Figure~\ref{fig:import}.
The annotation tool is designed to support previously annotated images with existing annotation in JSON files that are located in the same folder as the annotated images. Images uploaded with JSON files will have existing annotations automatically populated in the review table. The JSON file contains data structure format that is commonly used in annotation tools to store annotated labels and locations.
Our tool is designed to operate as a standalone system where images and annotation data can be uploaded from the local computer. It can be easily integrated into an existing image-based dietary assessment system where captured images are loaded directly from the system's database.

\begin{figure}[t]
    \centering
    \begin{subfigure}
        \centering
        \includegraphics[width=0.6\columnwidth]{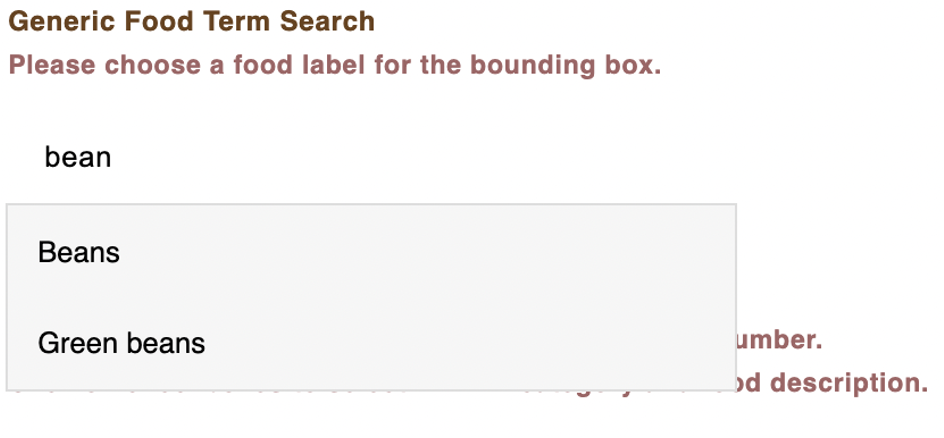}
        \caption{Example of search "bean" in the generic food term which is used for image recognition task}
        \label{fig:generic}
    \end{subfigure}
    % \vfill
    \begin{subfigure}
     \centering
     \includegraphics[width=0.6\columnwidth]{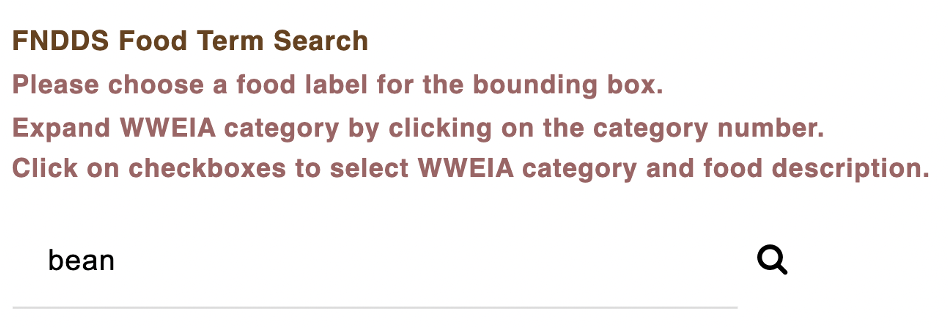}
     \caption{Example of search USDA food code and associated nutritional food term that contains "bean"}
     \label{fig:fndds}
    \end{subfigure}

    %  \caption{}
    %  \label{}
\end{figure}

\subsubsection{Food Item Localization}
Our tool provides an interactive design for food item localization, which allows researchers to modify or create a bounding box on the food image using the hold-and-drag operation. 
As shown in Figure~\ref{fig:loc}, the instructions for this step are displayed on the top to provide clear guideline to the researchers. 
Once the bounding box is drawn, a cropped image of the extracted region is shown in the review section for researchers to verify if the bounding box is drawn accurately. This step is particularly useful for complex images with multiple food items. 

\begin{table*}[t]
    \caption{Table~\ref{tab:ex}: Examples of images in the nutrition and food group based image database with the associated generic food labels and hierarchical food item name and associated sub-category and category confirmed by the dietitians}
    \label{tab:ex}
    \begin{center}
    \resizebox{0.75\textwidth}{!}{
    \begin{tabular}{ |p{0.7\columnwidth}||p{0.6\columnwidth}|p{0.6\columnwidth}| } 
     \hline
     Example Image & \includegraphics[width=0.127\textwidth]{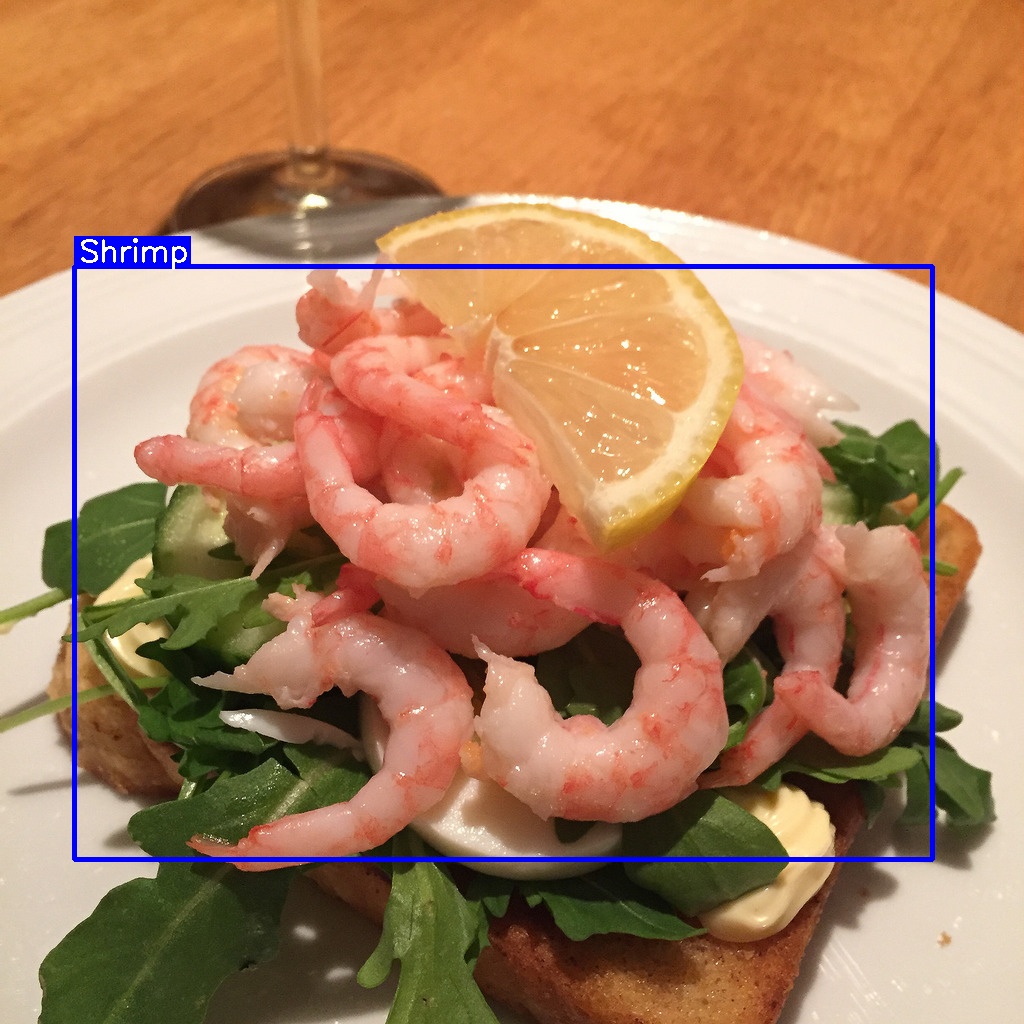} & \includegraphics[width=0.127\textwidth]{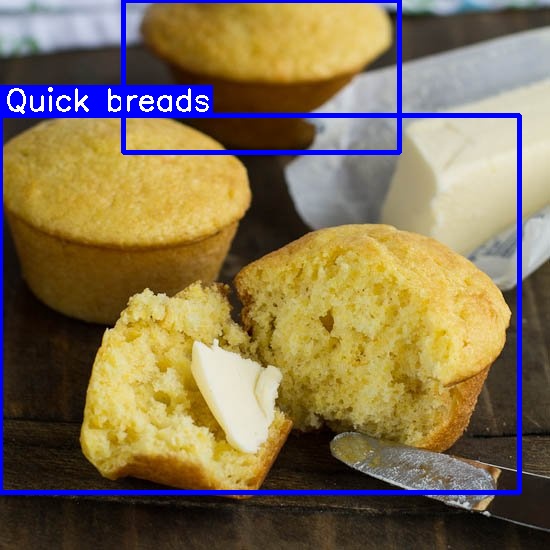} \\ 
     \hline
     Generic Food Label & Shrimp & Quick bread \\ 
     \hline
     WWEIA nutrition group category & Protein Foods & Grains \\ 
     \hline
     WWEIA nutrition subgroup category & Shellfish & Biscuits, muffins, quick breads \\ 
     \hline
     General USDA food code & 26319110 & 52201000 \\ 
     \hline
     General USDA food code description & Shrimp, cooked, NS as to cooking method & Cornbread, prepared from mix\\ 
     \hline
     Specific USDA food code & 26319180 & 52206010 \\ 
     \hline
     Specific USDA food code description & Shrimp, canned & Cornbread muffin, stick, round \\ 
     \hline
    \end{tabular}
    }
    \end{center}
\vspace{-1.0cm}
\end{table*}

\subsubsection{Nutrition Information Match}
Once the bounding box is confirmed, the researcher can select the WWEIA food sub-category and USDA food codes that match the cropped region of the image by using the search tools.
We designed two separate search tools. The first one is for generic food labels and it is used for the image recognition task (Figure~\ref{fig:generic}). The second search tool is for WWEIA food sub-category and USDA food codes annotation which is used for further nutrition analysis (Figure~\ref{fig:fndds}). The USDA food code and corresponding description are annotated with this search tool. 

\begin{figure}[t]
    \centering
    \begin{subfigure}
    \centering
     \includegraphics[width=0.6\columnwidth]{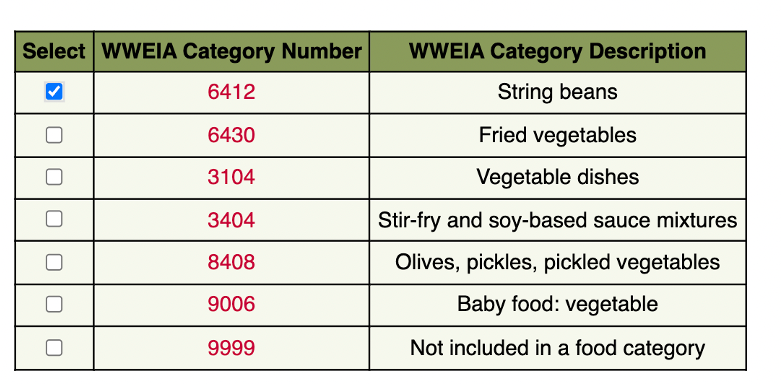}
     \caption{Example of WWEIA food category table shown if the search keyword is "bean"}
     \label{fig:wweiatab}
    \end{subfigure}
    \begin{subfigure}
         \centering
         \includegraphics[width=0.8\columnwidth]{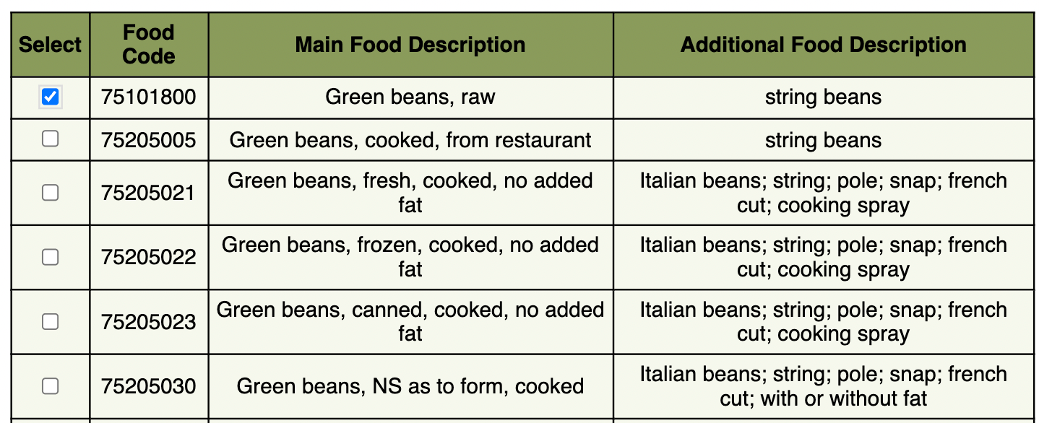}
         \caption{Example of USDA food code and associated food code description shown in a table if the "string bean" is selected as WWEIA food category}
         \label{fig:fnddstab}
    \end{subfigure}
    %  \caption{}
    %  \label{}
    \vspace{-0.5cm}
\end{figure}

\begin{figure}[t]
 \centering
 \includegraphics[width=0.85\columnwidth]{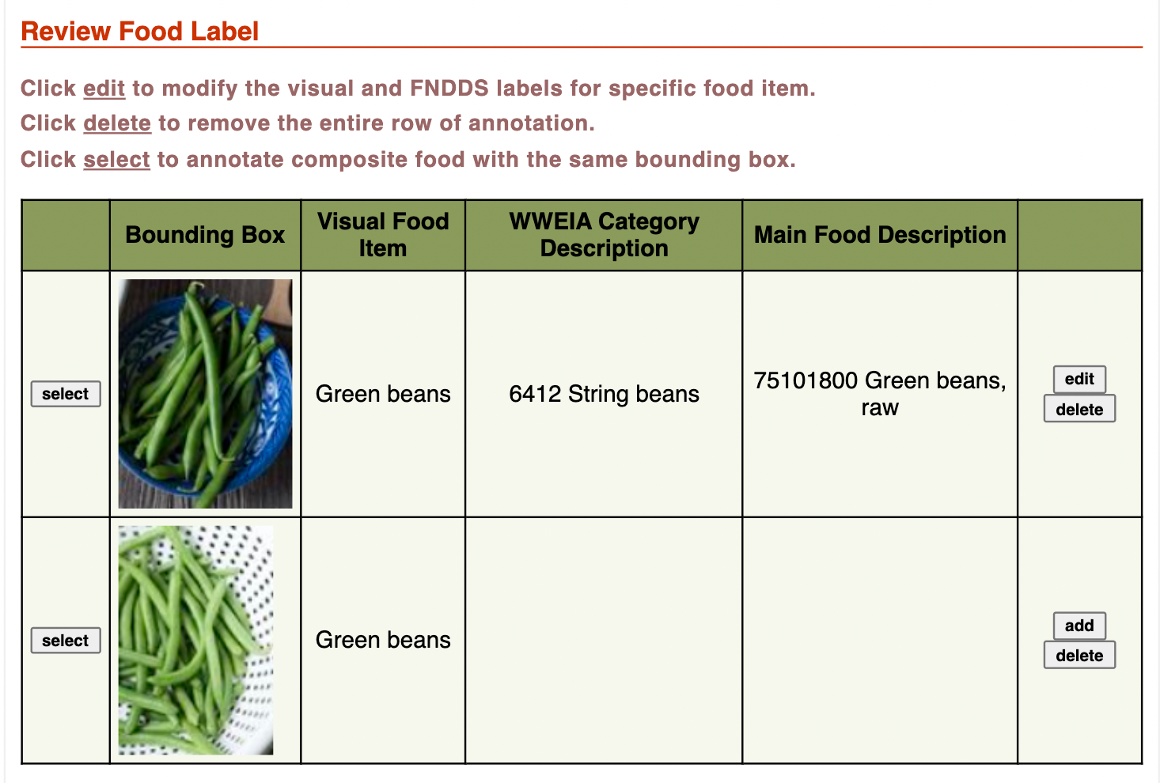}
 \caption{Screenshots of annotation review table}
 \label{fig:review}
 \vspace{-0.5cm}
\end{figure}

The WWEIA food sub-category and USDA food codes searching is implemented based on the hierarchical data structure of the WWEIA \cite{usdatool}.
Based on the search keyword, the relevant WWEIA food categories are listed in a table. 
An example table is shown in Figure~\ref{fig:wweiatab} for the search keyword ``bean".
Once we choose a WWEIA category from the table, the corresponding USDA food codes and associated food code descriptions are shown in a separate table. 
As shown in Figure~\ref{fig:fnddstab}, a table is populated if ``string bean" is selected.
Finally, a specific food code or multiple food codes can be selected and assigned to the chosen bounding box.

\subsubsection{Review of Annotation Results}
The annotated food labels and bounding boxes are loaded and displayed in the review table for researchers to review and edit, as shown in Figure~\ref{fig:review}.
We also included a download button to save the annotation results onto a local computer as JSON files.
When the annotation tool is integrated into an existing database system, the annotation results are managed by the database system and stored on the server end.

%%%%%%%%%%%%%%%%%%%%%%%%%%%%%%%%%%
% Results and Discussion
%%%%%%%%%%%%%%%%%%%%%%%%%%%%%%%%%%

\section{Results and Discussion}
The proposed method was used to build a nutrition and food group based image database containing the most frequently consumed food sub-categories and those contributing most to total energy of US adults in the United States.
We identified 74 WWEIA food sub-categories as commonly consumed ~\cite{lin2022nhames} and integrated them into the VIPER-FoodNet  (VFN) food image database~\cite{Mao2021} where the food images were downloaded from the Internet.
Each WWEIA food sub-category was assigned a general USDA food code, \textit{e.g.}, the USDA food code 42100100 and its corresponding description ``Almonds, non-specified" is assigned for the ``Almond" sub-category.
Each image with a WWEIA food sub-category was then reviewed by the nutrition researchers to determine if a more specific USDA food code (\textit{e.g.}, ``Almonds, salted" if visibly salt on the almond) should be assigned to the image considering the visual appearance of food item.
In total, there were 16,114 food images matched to 1,865 USDA food codes.
Examples of food images and their associated nutrition information are shown in Table~\ref{tab:ex}.

However, even nutrition researchers cannot always accurately recognize every food item in the image due to image resolution and factors that may mask or cover food including sauce and cooking methods and other nutrition related factofs. For example, it is not possible to identify what kind of fruit or meat is included in a pie by only looking at the outside appearance , not to mention occurrences where foods are visually similar but differ in nutrition content like lean or full fat beef.
% However, the online image cannot be guaranteed to visually recognize every food items, visually similar images may differ substantially in nutritional content.
Our ongoing work will focus on providing additional contextual information, such as recipe and ingredients, along with images for nutrition researchers to link the food item in the image to the USDA food code.

\section{Conclusion}
In this paper, we proposed to create a nutrition and food group based image database that may be linked to a nutrient database.
We designed a protocol to link USDA food codes to associated food images in a systematic, transparent, efficient way. 
We enhanced a previously developed annotation tool to provide new mechanisms for nutrition researchers to construct such an image database by considering both visual and food group based information.
We deploy the proposed framework to create a nutrition and food group based image database focusing on the most frequently consumed foods and those contributing most to energy intake in the United States, which will pave the way to address the lack of high quality food image datasets needed for deep learning based food image analysis.
% \section{Acknowledgments} 

% To start a new column (but not a new page) and help balance the last-page
% column length use \vfill\pagebreak.

%%%%%%%%%%%%%%%%%%%%%%%%%%%%%%%%%%
% Bibliography
%%%%%%%%%%%%%%%%%%%%%%%%%%%%%%%%%%

\small
\bibliographystyle{plain}
\bibliography{main}

%%%%%%%%%%%%%%%%%%%%%%%%%%%%%%%%%%
% Biography
%%%%%%%%%%%%%%%%%%%%%%%%%%%%%%%%%%

\begin{biography}
\textbf{Zeman Shao} received his B.S. degree in Software Engineering from Tongji University, Shanghai, China in 2016. He is currently working toward the Ph.D degree in Electrical and Computer Engineering at Purdue University, West Lafayette, IN, USA. His research interests include image processing, computer vision and deep learning.

\textbf{Jiangpeng He} received his B.S. degree in Electrical and Electronic Engineering from University of Electronic Science and Technology of China in July 2017. He is currently a Ph.D. student at the School of Electrical and Computer Engineering, Purdue University, West Lafayette, IN, USA. His research interests include image processing, computer vision and deep learning.

\textbf{Ya-Yuan (Tiffany) Yu} is pursuing her BS in electrical engineering from Purdue University and is expected to graduate in 2023. She worked as a research assistant for the Technology Assisted Dietary Assessment (TADA) project at Purdue University. She also joined a food detection project team during the school years. Her fields of focus are image processing, communications, and data analysis. She is a member of Eta Kappa Nu Beta chapter and IEEE. 

\textbf{Luotao Lin} achieved his Bachelor of Medicine in preventive medicine (food hygiene and nutrition direction) at Chongqing Medical University in China and Master of Science in nutrition science at University of Massachusetts Amherst. Currently, he is a PhD candidate of Department of Nutrition Science at Purdue in West Lafayette, IN and focuses on nutrition epidemiology including temporal lifestyle behavior pattern and precise nutrition. He is currently a member of the American Society of Nutrition.

\textbf{Alexandra E. Cowan} received her PhD in Nutrition Science from Purdue University (2022). She is currently a Graduate Research Assistant in the Department of Nutrition Science at Purdue University in West Lafayette, IN. She is a nutritional epidemiologist focused on improving and developing new methods of dietary assessment, assessing total micronutrient exposures of the U.S. population, and dietary supplement research. She is currently a member of the American Society of Nutrition. 

\textbf{Heather A. Eicher-Miller} received her PhD in Foods and Nutrition from Purdue University (2009). She is currently an Associate Professor in the Department of Nutrition Science at Purdue University in West Lafayette, IN. She is a nutrition epidemiologist focused on developing new methods of dietary assessment, discovery and evaluation of dietary and lifestyle patterns, and improving food security among low-resource populations. She is a member of the American Society of Nutrition and the Society for Nutrition Education and Behavior and a member of the Board of Editors for the “Journal of the Academy of Nutrition and Dietetics” and the journal, “Advances in Nutrition”.

\textbf{Fengqing Zhu} is an Assistant Professor of Electrical and Computer Engineering at Purdue University, West Lafayette, Indiana. Dr. Zhu received the B.S.E.E. (with highest distinction), M.S. and Ph.D. degrees in Electrical and Computer Engineering from Purdue University in 2004, 2006 and 2011, respectively. Her research interests include image processing, computer vision, video compression and digital health. Prior to joining Purdue in 2015, she was a Staff Researcher at Futurewei Technologies (USA). She is a senior member of the IEEE.

\end{biography}

\end{document}